\documentclass{article}

     \PassOptionsToPackage{numbers, compress}{natbib}
    
\usepackage[preprint]{neurips_data_2024}
\usepackage{wrapfig}

\usepackage[utf8]{inputenc} 
\usepackage[T1]{fontenc}    
\usepackage{hyperref}       
\usepackage{url}            
\usepackage{booktabs}       
\usepackage{amsfonts}       
\usepackage{nicefrac}       
\usepackage{microtype}      
\usepackage{xcolor}         
\usepackage{graphicx}
\usepackage{listings}
\usepackage{enumitem}
\usepackage{subcaption}
\usepackage{wrapfig,lipsum,}

\newcommand{\ours}{PARSE-Ego4D}

\title{PARSE-Ego4D: Personal Action Recommendation Suggestions for Egocentric Videos}

\author{%
Steven Abreu$^{1,2}$ \quad Tiffany D. Do$^{1,3}$ \quad Karan Ahuja$^1$ \quad Eric J. Gonzalez$^1$ \\ 
\textbf{Lee Payne$^1$ \quad Daniel McDuff $^1$ \quad Mar Gonzalez-Franco $^1$} \\
$^1$Google \quad $^2$University of Groningen\quad $^3$University of Central Florida\\
\texttt{s.abreu@rug.nl, tiffany.do@ucf.edu}, \\
\texttt{\{karanahuja,ejgonz,leepayne,dmcduff,margon\}@google.com}
}

\begin{document}

\maketitle

\vspace{-0.8cm}

\begin{figure}[h!]
    \centering
    \includegraphics[width=0.9\textwidth]{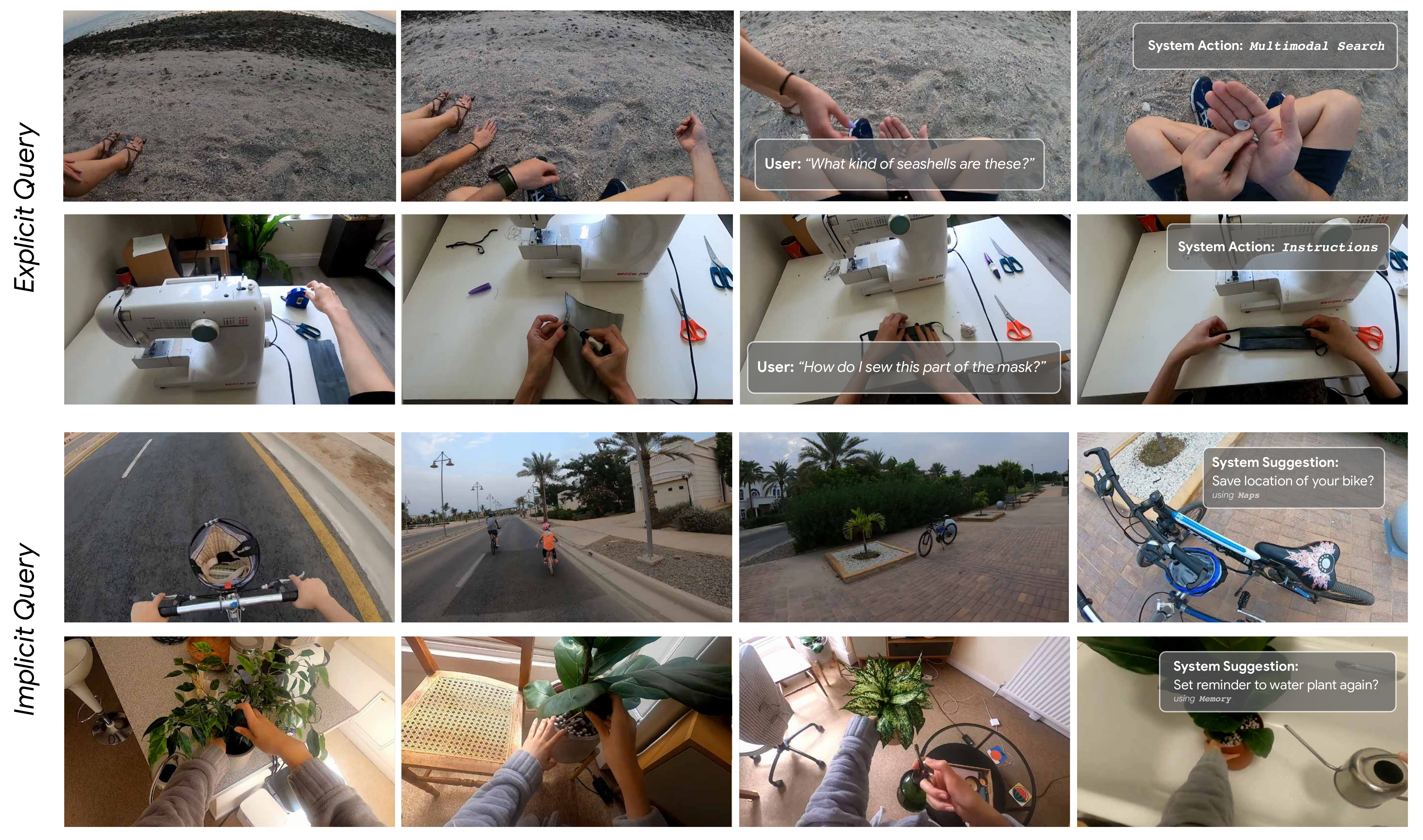}
    \caption{
    Examples of action suggestions for different videos in the \ours{} dataset. 
    }
    \label{fig:teaser}
\end{figure}

\vspace{-0.3cm}

\begin{abstract}
Intelligent assistance involves not only understanding but also action. Existing ego-centric video datasets contain rich annotations of the videos, but not of actions that an intelligent assistant could perform in the moment. To address this gap, we release \textbf{\ours{}}, a new set of personal action recommendation annotations for the Ego4D dataset. We take a multi-stage approach to generating and evaluating these annotations. 
First, we used a prompt-engineered large language model (LLM) to generate context-aware action suggestions and identified over 18,000 action suggestions. While these synthetic action suggestions are valuable, the inherent limitations of LLMs necessitate human evaluation. To ensure high-quality and user-centered recommendations, we conducted a large-scale human annotation study that provides grounding in human preferences for all of \ours{}. 
We analyze the inter-rater agreement and evaluate subjective preferences of participants.
Based on our synthetic dataset and complete human annotations, we propose several new tasks for action suggestions based on ego-centric videos. 
We encourage novel solutions that improve latency and energy requirements. 
The annotations in \ours{} will support researchers and developers who are working on building action recommendation systems for augmented and virtual reality systems.
\end{abstract}


\section{Introduction}

Egocentric perception, the ability to capture and understanding of the world from a first-person perspective is gaining significant traction with the adoption of Augmented Reality (AR) and Head-Mounted Displays. Recent advancements in egocentric video understanding have opened new opportunities for research and application, including activity recognition \cite{bohus2021platform,Liu2024Human}, object interaction analysis \cite{dogan2024augmented, bohus2024sigma, wang2023holoassist}, and social interaction modeling \cite{huang2024egoexolearn}. However, a fundamental limitation of most existing systems is their reactive nature, driven by explicit user queries. We argue that the ability to take bespoke, proactive actions that anticipate a user's needs is a core component of intelligent behavior without which these systems will be limited in their practical applications.



Public datasets have been highly consequential in the advancement of machine learning and artificial intelligence. However, older datasets, particularly in the field of computer vision, often included static, context agnostic, unimodal repositories of labeled data, \textit{e.g.}, COCO~\cite{lin2014coco} or Imagenet~\cite{russakovsky2015imagenet}. As ambitions in AI have become more complex and situated in the context of specific human-computer interaction scenarios, there has been a movement toward datasets that contain temporal, ecologically valid and multimodal data. This paradigm shift is exemplified in new datasets such as Ego4D \cite{grauman2022ego4d} or Ego-Exo4D \cite{grauman2023ego} which include thousands of hours of egocentric video streams.
Several existing egocentric vision datasets provide rich annotations for tasks like activity recognition \cite{chung2023enabling,egotaskqa2022,openvocab2023,ego4dgoalstep2023,actionsense2022}, object tracking \cite{egotracks2023}, and for the analysis of interactions with other humans \cite{conflab2022} and with the environment \cite{chang2023lookma,egoenv2023}. These datasets play a crucial role in advancing research on egocentric perception.
However, previous work focucses primarily on understanding and classifying video content. While valuable, such annotations don't address how an intelligent system could suggest and take actions in the real or virtual world to assist the user. 
This ability to take appropriate action is a core component of intelligent behavior. Without this capability, systems can simply observe the world but have limited practical application as they rely on explicit user queries, as in existing work in visual question answering \cite{Fan_2019_ICCV} and visual query localization \cite{egovql2023}. The ability to generate bespoke or proactive actions, which could further our exploration of the environment, is currently missing.

To address this limitation and empower the development of proactive AI assistants, we release \textbf{\ours{}}, a novel dataset designed to provide personal action recommendation annotations for egocentric videos. Herein, we consider personal suggestions that are context-dependent \cite{ghiani2017personalization}.
Our dataset is built upon the extensive Ego4D dataset~\cite{grauman2022ego4d}, which contains 3,670 hours of first-person video recordings of a wide range of everyday activities. 
We leverage a two-stage annotation process, combining automated suggestions generated by a state-of-the-art large language model (Gemini Pro \cite{team2023gemini}) with meticulous human evaluation, to ensure the quality, relevance, and usefulness of the action recommendations. These annotations identify moments in the Ego4D video sequence when an assistant may be able to suggest a useful action (see more details in Section \ref{sec:dataset}), creating a total of 18,360 possible action recommendations, which we call the \emph{synthetic} dataset for it was created by an LLM and not yet grounded in human preferences. While the AI-assisted nature of these annotations allowed us to generate them at scale, the quality can be called into question. Consequently, we performed a large-scale human validation study that provides the necessary grounding in human preferences.

Using a 5-point Likert scale for human ratings, we found that 65\% of all synthetically generated action suggestions were annotated with average scores above 3, and 42\% were annotated with average scores above 4. Considering that our dataset aims at providing a footing to fine tune existing agents so they can provide better actions and personalized queries on-the-fly using real-time multi-modal data, the relatively high scoring validates our automatic captioning and annotation approach.
 
Our first study took 20 samples from our newly generated \ours{} dataset and requested 20 human participants to evaluate our AI-generated queries and action suggestions with respect to five axes: (1) whether the query was \texttt{sensible} at all (to filter out hallucinations and mistakes from the LLM), (2) whether the suggestion would be helpful as an \texttt{implicit} suggestion if it was presented unsolicited to the user, (3) whether the action suggestion was \texttt{valuable} to the user (\textit{e.g.}, by saving them time), (4) whether the suggested action was the \texttt{correct} action to take in response to the query, and (5) if the participant would personally be \texttt{likely} to take the presented action on their AR glasses (see Figure \ref{fig:survey-design}).
In the large-scale annotation study, we requested 20\% of the \ours{} dataset to be annotated by 5 human raters, and the remaining 80\% of the \ours{} dataset to be annotated by 1 human rater. For the annotation study, we only evaluated the (1) sensibleness, (2) the helpfulness as an implicit action suggestion, and (3) the correctness of the action. 

The current \textbf{\ours{}}  dataset aims at providing a basis for fine-tuning existing agents so they can provide better actions and queries on the fly using real-time multimodal data. Annotation, code and model responses can be found at: \url{https://parse-ego4d.github.io}.



\section{Related Work}

\paragraph{Human Computer Interaction}

Within the realm of Human-Computer Interaction (HCI), research on action recommendations has primarily focused on enhancing user experience and task efficiency \cite{amershi2019guidelines}. Prior work has identified several key motivations for providing action suggestions in user interfaces (UIs): saving time by streamlining interactions \cite{dogan2024augmented,wang2023holoassist}, improving discoverability of features and functionalities \cite{suh2023sensecape, huang2024egoexolearn}, and enabling discrete interactions without explicit user input \cite{vogel2004interactive,schmidt2000implicit} -- an aspect that is particularly relevant for AR glasses.

Research on spatial UI transitions in AR has explored the balance between automation and user control in placing and manipulating UI elements \cite{lu2022exploring}, emphasizing the importance of user agency and control for a positive user experience. This underscores the need for easy error recovery mechanisms to mitigate the negative impact of incorrect predictions or actions.
Explainability has emerged as a crucial aspect of action recommendations, particularly in the context of augmented reality (AR) systems. Xu et al. \cite{xu2023xair} introduced the XAIR framework, emphasizing the importance of providing clear and understandable explanations for AI-generated suggestions in AR environments. Their findings highlight that users prefer personalized explanations and that the timing, content, and modality of explanations should be carefully tailored to the user's context and goals.

\paragraph{Machine Learning}

The increasing traction of egocentric devices 
through smart glasses, like Snap's Spectacles \cite{snapspectacles} and Meta's Ray-Ban Stories \cite{metaraybanstories}, and mixed reality head-mounted displays, like Apple's Vision Pro \cite{applevisionpro} and Meta's Quest \cite{metaquest}, 
has spurred significant advancements in egocentric video \cite{grauman2022ego4d} and user understanding \cite{grauman2023ego, song2024ego4d}. These devices provide a unique perspective on the user's environment and activities, making them ideal platforms for personalized and context-aware AI assistants. The recent surge in multi-modal Large Language Models (M-LLMs) such as Gemini \cite{team2023gemini} and ChatGPT \cite{chatgpt} has further propelled research in this area, particularly in the realm of visual perception and question answering. In the realm of egocentric video understanding, works like EgoOnly \cite{wang2023ego} have explored action detection without relying on exocentric (third-person) data, demonstrating the potential of understanding actions from a first-person perspective as a prerequisite for generating relevant action suggestions. Additionally, research in intent classification, such as IntentCapsNet \cite{xia2018zero}, aims to discern user needs and preferences from egocentric videos, which can inform the generation of personalized suggestions.

Recent research has also focused on developing agents that can understand and execute instructions in interactive environments. In robotics, Instruct2Act \cite{huang2023instruct2act} leverages LLMs to generate code that controls a robotic arm to manipulate objects based on multi-modal instructions. In UI interaction, approaches like CogAgent \cite{hong2023cogagent} have shown promising results in mapping natural language instructions to sequences of actions on mobile devices. Similarly, a plethora of LLM-based action agents are aiding in tasks such as knowledge discovery \cite{nakano2021webgpt}, web navigation \cite{liu2023alltogether}, and shopping \cite{yao2022webshop}, among others.

Despite these advancements in understanding actions and executing instructions, there remains a gap in the development of proactive AI assistants for egocentric devices. Existing datasets like Ego4D \cite{grauman2022ego4d} and EPIC-Kitchens \cite{damen2020epic} provide rich annotations for understanding activities and objects but do not offer a direct mapping to actionable recommendations. 

The form factor and resource limitations of AR/VR devices, impose unique challenges on the machine learning models used in these systems.  Energy efficiency, latency, and memory footprint are critical concerns for ensuring a positive user experience in these battery-powered and often mobile environments. Lightweight LLM models like Gemini XXS \cite{team2023gemini} are moving towards deployment on resource-constrained devices. Moreover, model compression techniques like quantization \cite{hubrara2018quantized} have been applied to transformer architectures \cite{wang2023bitnet,ma2024era} as well as pruning \cite{llmpruner2023}. Furthermore, more efficient architectures are being developed that compete with transformers and offer better scaling with sequence length \cite{botev2024recurrentgemma,gu2023mamba,dao2024transformers}. Model compression techniques and novel architectures for sequence modeling may provide a path towards efficient always-on foundation models running on resource-constrained AR/VR devices.

\section{The \ours{} Dataset}  
\label{sec:dataset}

The \ours{} dataset builds on top of the Ego4D dataset \cite{grauman2022ego4d} and provides action suggestions that draw from the specification of available actions given in Section \ref{ss:available-actions}. After generating synthetic action suggestions using an LLM (Section \ref{ss:synthetic-llm-annotation}), all action suggestions are rated through in a human annotation study (Section \ref{ss:human-annotation-study}).

\subsection{The Ego4D dataset}

The Ego4D dataset is a massive ego-centric video dataset containing 3,670 hours of daily-life activity video from over 900 people across 74 locations and 9 countries. The data is split into $\approx$9,600 videos with an average duration of 15-30 minutes and contains video streams from a head-mounted camera, as well as IMU and gaze data.
The Ego4D dataset further contains rich annotations. All videos have dense written narrations in English for intervals of $\approx$10 seconds, as well as a summary for the whole video clip. Additionally, transcriptions, speech segmentation, user attention, speech target classification, speaker labeling, and episodic memory annotations are also provided for parts, or all, of the Ego4D dataset. We make use of the egocentric videos as well as the complete textual narrations from the Ego4D dataset.

Adding additional annotations and expanding the utility of such a dataset that already been collected is better than collecting a new dataset for two reasons. 
\textbf{(1)} It enables us to focus on the action suggestions without having to dedicate additional compute to labeling the narrations and captioning and labeling a whole new dataset.
\textbf{(2)} Given the substantial investment made into this dataset, we can build on top of other projects that also have augmented the existing Ego4D \cite{ego4dgoalstep2023,egotracks2023}.

\subsection{Available actions}
\label{ss:available-actions}

To create a dataset with action suggestions, we first identify a set of possible actions that can be invoked from the AR/VR device, considering applications that future AR/VR devices are expected to support, such as:
\begin{itemize}[leftmargin=*]
    \item \textbf{Search}: an application that can take in the current camera input and a query (written or spoken) to run a multimodal search, and provide a written and/or spoken response.
    \item \textbf{Assistant search}: the AI assistant for the device, with access to system apps like ``notes'', ``timer'', ``stopwatch'', ``alarm'', ``email'', ``music'', ``phone'', ``contacts'', ``messages'', ``settings'', ``calculator'' and potentially more such as smart home access, notification access, and more. 
    \item \textbf{Assistant local}: an application that can explicitly store memories and retrieve them later. Memories may be enrolled manually and explicitly, but they may also be enrolled passively and automatically as in the episodic memory tasks from the Ego4D dataset \cite{grauman2022ego4d}.
    \item \textbf{Language}: an application that can either transcribe what the user is hearing right now, translate what the user is reading or hearing, or determine what language is spoken.
    \item \textbf{Directions}: find relevant places nearby, plan routes, estimate distances and navigate to places.
    \item \textbf{Assistant guide}: an application that can give detailed and step-by-step instructions to the user.
    \item \textbf{Others}: For open-ended exploration, we also define the option to suggest actions that do not belong to the categories mentioned above. This may allow the LLM to come up with novel, creative use cases for AR glasses that are not covered by the available applications listed above. Actions that fall into this category are not included in the human annotation study.
\end{itemize}

\begin{figure*}[t]
    \centering
    \includegraphics[width=1.0\textwidth]{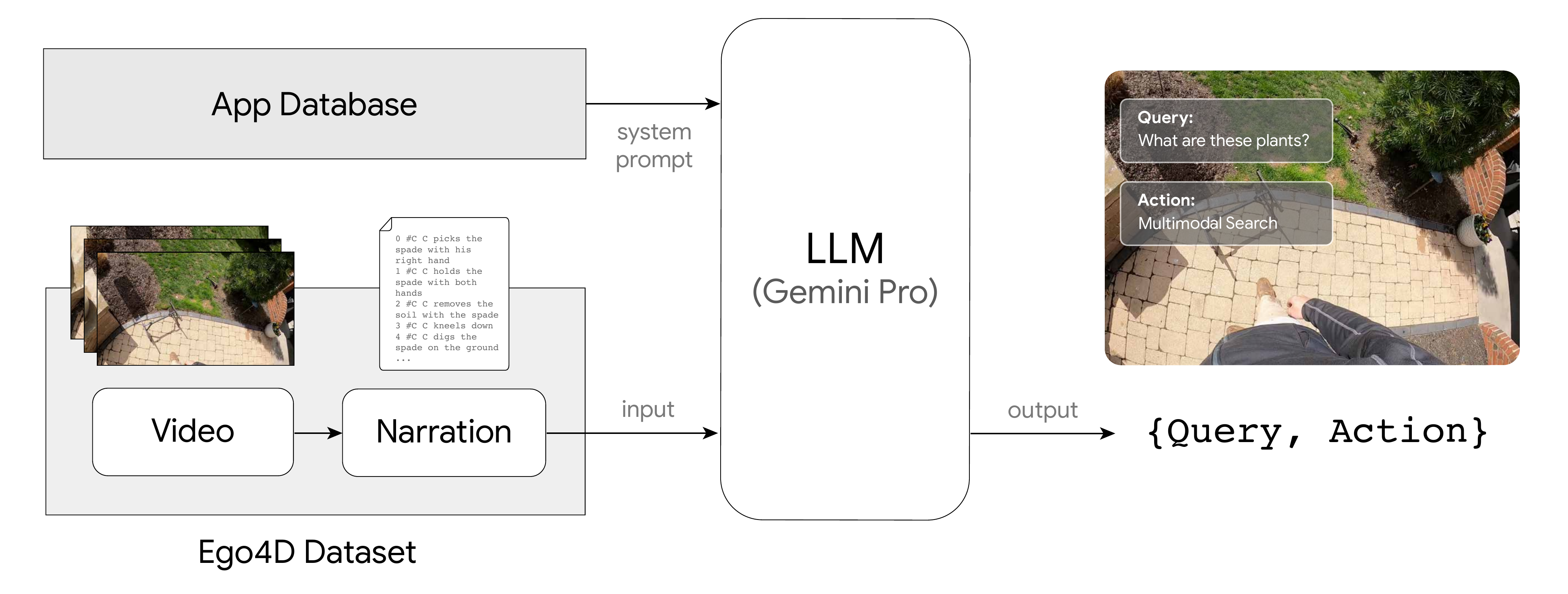}
    \caption{\textbf{\ours{}} - We curated, annotated and open-source over 11,000 action suggestions for the Ego4D dataset. These annotations support researchers and developers who are working on building personalized action recommendation systems for augmented and virtual reality systems.}
    \label{fig:synthetic-annotation}
    \vspace{-0.5em}
\end{figure*}

\subsection{Synthetic LLM annotation}
\label{ss:synthetic-llm-annotation}

In order to generate samples for action suggestions we used a prompt-engineered LLM, the Gemini Pro model \cite{team2023gemini}. We use prompt engineering for the LLM to use in-context learning to learn the annotation task. We pass textual narration sentences from the Ego4D annotations as input to the LLM, and request a JSON-formatted output in response. The process is illustrated in Figure \ref{fig:synthetic-annotation}.
The system prompt to the LLM contains:
\begin{itemize}[leftmargin=*]
    \item \textbf{Task explanation}: the LLM is prompted to behave as a user experience researcher, helping to collect a dataset for useful interactions with AR glasses.
    \item \textbf{Input format}: the input format of the narrations is explained and an example is presented.
    \item \textbf{Available actions}: the set of available actions described in Section \ref{ss:available-actions} is listed with example queries and the expected API format (this API format is not used for the annotation study).
    \item \textbf{Output format}: the expected JSON output format is described. The LLM is expected to return its \texttt{thoughts} to assess the situation and develop a rationale for the suggestion that it will return, the \texttt{query} that the user would ask along with the timestamp when this would be asked, and the corresponding \texttt{action} that the system should take in response to the query. 
\end{itemize}
For every video clip in the Ego4D dataset, we split the entire video into batches of 200 narration sentences and pass these batches into the LLM. We drop 1897 short videos that have fewer than 50 sentences of narrations and do not generate any action suggestions for these.
If the response of the LLM is not in valid JSON format, we ask the LLM to re-generate it to be valid. Once the LLM has generated a valid suggestion, we ask it to generate one more suggestion for the same input data. The complete system prompt is given in the Supplementary Materials.

\begin{figure}[h]
    \centering
    \includegraphics[width=\textwidth]{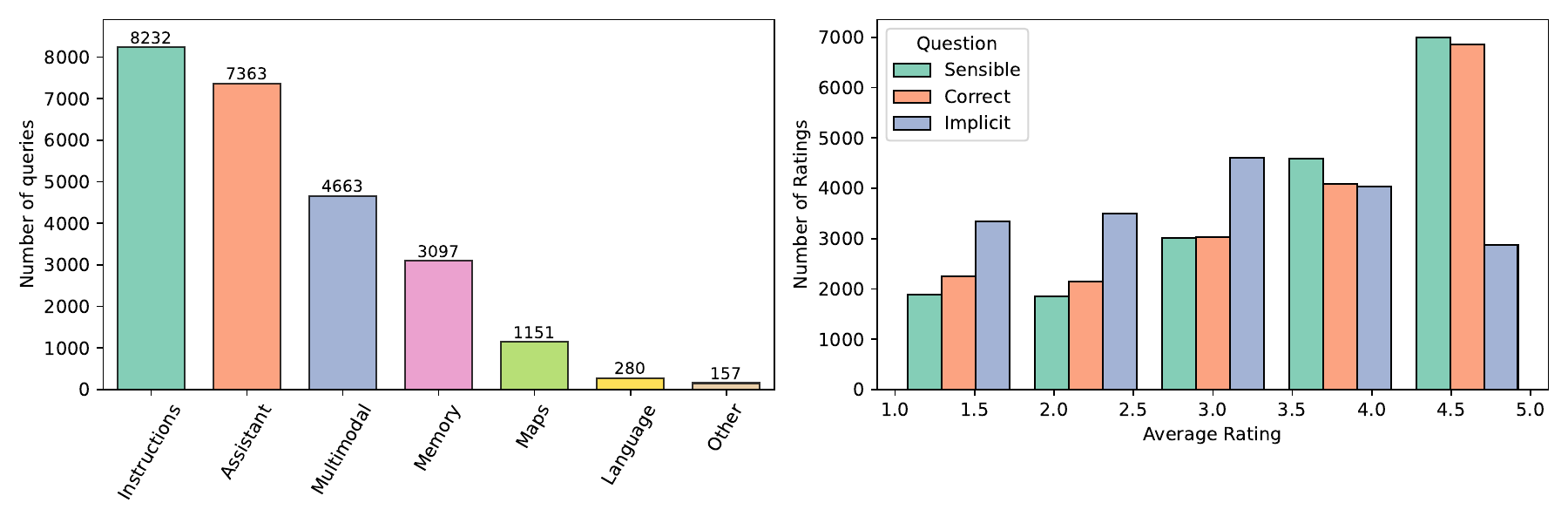}
    \caption{
    \textbf{Left}: Suggested actions by type. \textbf{Right}: Score distribution for different questions in the human annotation study, showing that there are more valid explicit suggestions than implicit suggestions.
    }
    \label{fig:scores}
    \label{fig:suggestedresults}
\end{figure}

The resulting dataset of synthetically generated action suggestions contains 32,155 action suggestions. After removing 7,491 duplicates (where the same batch of narrations gives the same query and action), we also remove 2,575 approximate duplicates. We classify a suggestion to be an approximate duplicate if it has an embedding distance $f(x_1,x_2)>0.9$ using the normalized Gemini text embeddings from the Gemini API\footnote{\href{https://ai.google.dev/gemini-api/docs/models/gemini\#text-embedding}{ai.google.dev/gemini-api/docs/models/gemini\#text-embedding}}. This leaves 19,255 suggestions in our synthetic dataset, see Figure \ref{fig:suggestedresults} (left).

Every sample in the \ours{} dataset contains 
a reference to the Ego4D video, 
a time range that corresponds to the narration sentence during which the action suggestion is invoked, 
the suggestion in the form of a (query, action) tuple, 
the name of the LLM that was used to generate the suggesion.
Additionally, each sample also contains a parameter JSON that provides structured information that the suggested application may use. 
Furthermore, the dataset contains a rationale for each sample that was generated by the LLM as a form of chain-of-thought reasoning \cite{wei2022chain}. 
We do not include the action parameters or rationale in the human annotation study, but still provide them as part of the \ours{} dataset.

\subsection{Human annotation study}
\label{ss:human-annotation-study}

\begin{figure}[t]
    \centering
    \includegraphics[width=\textwidth]{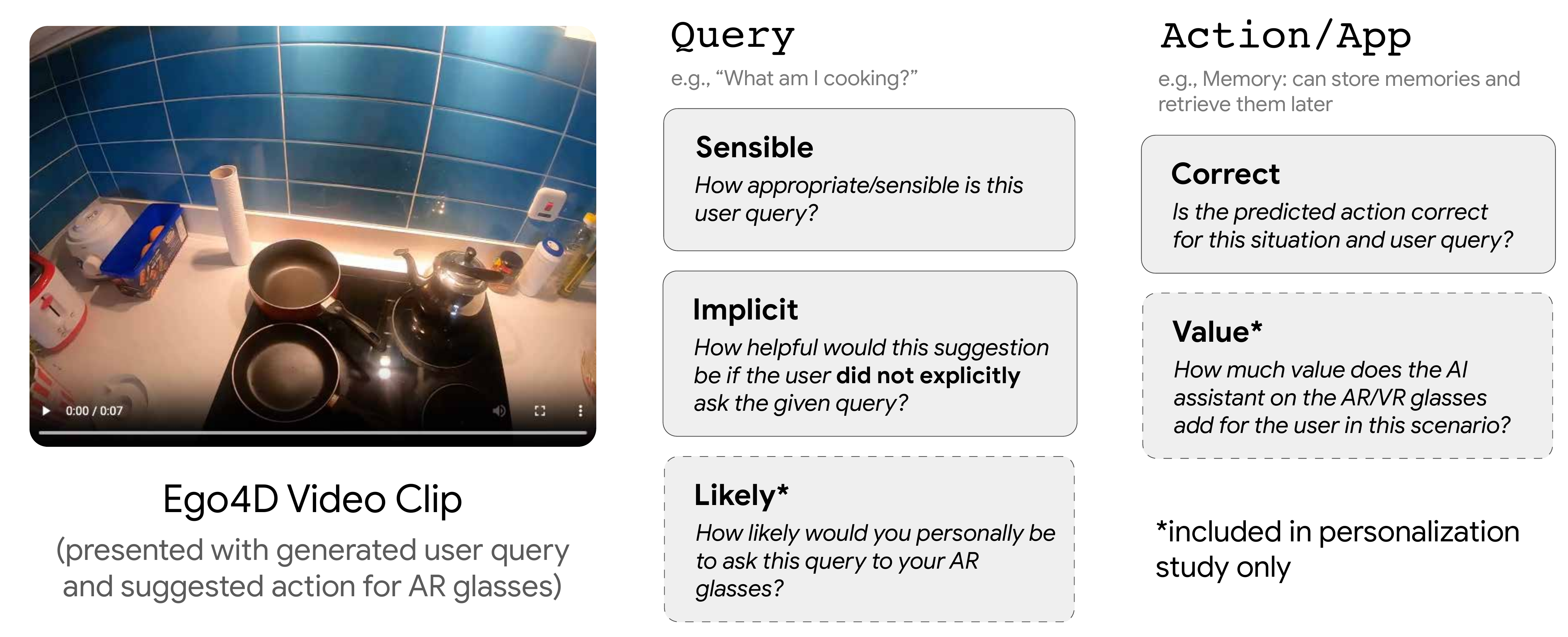}
    \caption{Sketch of the survey that participants filled out in the human annotation study in order to verify the synthetically generated action suggestions in \ours{}.}
    \label{fig:survey-design}
    \vspace{-1em}
\end{figure}

We annotate 20\% of the synthetic action suggestion dataset gathered in Section \ref{ss:synthetic-llm-annotation} with 5 human raters which will be used as the test split. We annotate the remaining 80\% of the dataset with 1 human rater each--of which 75\% will be used as the train set and the other 5\% as the validation set. In total, we received 36,171 annotations for 18,360 suggestions.
The originally published benchmarks for the Ego4D dataset come with several different train/test/validation splits. However, these data splits are either based on subsets of the entire dataset, or based on specific scenarios, \textit{e.g.}, hand-object interactions. As we are using the entirety of the Ego4D dataset, we chose a new random train/test/validation split.

The survey for participants of the annotation study is shown in Figure \ref{fig:survey-design}. In the large-scale annotation study, each sample is evaluated with three separate questions that each verify one dimension of the \ours{} dataset. First, the sample is evaluated on being \texttt{sensible} to verify that the query makes sense in the given context. Second, query is being evaluated on being helpful as an \texttt{implicit} (or proactive) action suggestion. We expect that not all samples that score high on the \texttt{sensible} rating will also score highly on the \texttt{implicit} rating because we would expect users to have higher standards for implicit, proactive suggestions where false positives are disturbing or even annoying. Indeed, results from our annotation study confirm this, see Figure \ref{fig:suggestedresults}. Third, the action is evaluated for being \texttt{correct} given the query and context.

The release the \ours{} dataset with all suggestions and their corresponding ratings from human annotators. For all downstream experiments, we filter the dataset to keep only suggestions that have (mean) ratings \texttt{sensible >= 4} and \texttt{correct >= 4} to use only verified, high-quality suggestions. If only the queries are used and actions are discarded, we suggest filtering for \texttt{sensible >= 4}. For implicit, proactive suggestions we additionally filter for \texttt{implicit >= 4}. Optionally, the cutoff for mean ratings can also be set at $\mu=3$ instead of $\mu=4$.

\subsection{Subjective user study}

In addition to providing annotations to verify and ground our synthetic action suggestions in human preferences, we ran two extended surveys for participants to assess their subjective preferences for different action suggestions. We ran one study with $N=10$ participants and $M=10$ samples, and one study with $N=20$ participants and $M=20$ samples per participant. In these smaller subjective user studies, each participant is requested to answer all questions from the annotation survey shown in Figure \ref{fig:survey-design}.
In addition to the questions outlined in the previous section, participants of the subjective user study were also asked to evaluate how \texttt{likely} they would personally be to ask the given query to their AR glasses, and how much \texttt{value} they think an AI assistant would add in the given scenario. 

\begin{wraptable}{l}{4cm}
\centering
\caption{Intraclass Correlation Coefficients (ICC) for the Annotation Questions.}\label{tab:icc}
\begin{tabular}{rl}\\\toprule[1.5pt]
\textbf{Rating} & \textbf{ICC} \\\midrule
Sensible & 0.87\\
Helpful & 0.73\\
Value & 0.88\\
Likely & 0.90\\
Correct & 0.81\\
\bottomrule[1.5pt]
\end{tabular}
\end{wraptable} 

With these questions, we aim to better understand what kind of interactions different users value and to assess if there is a need for personalization in action recommendation systems based on our proposed action specification. 
Our results show that intraclass correlation coefficients (ICC) for the five annotation questions were above 0.7 for all questions and above 0.8 for all non-subjective questions from the study, thus showing high inter-rater agreement (see Table~\ref{tab:icc}).

Although the ICC for the personal \texttt{helpful} question is lower that for other questions, the inter-rater agreement is still considerably high. We thus conclude that personalization may not be very important for building useful and valuable action recommendation systems of the sort that are described in this paper. However, we acknowledge that our user study was small and that the actions used in the annotations studies do not allow for the kind of personal data to be used that would be available to a real-world assistant on augmented and virtual reality systems. We hypothesize that expanding the set of available actions and giving the AI assistant access to personal user data would strengthen the need for personalization in action suggestion systems.

\subsection{Participants} 
Participants for both the subjective and annotation studies were recruited from Prolific, an online platform for crowdworkers, and were pre-screened for English fluency. For the larger subjective user study, we recruited 20 participants (10 male, 10 female) with an average age of 27.47 (SD=7.80). Participants were geographically diverse, residing in Poland (7), Portugal (6), Hungary (2), South Africa (2), Germany (1), Italy (1), Spain (1), and New Zealand (1).

The annotation study involved 1496 participants (749 male, 747 female), with an average age of 29.83 (SD=9.15). Figure \ref{fig:demographics} presents a demographic breakdown of our participants, including gender, race, age, and country of residence. Participants annotated up to 20 samples each and were compensated through Prolific with US\$0.13 per annotation for an average hourly wage of US\$8.79.

\section{The \ours{} Benchmark}

We propose two tasks for action recommendation based on the \ours{} dataset. Each task aims to build action recommendation systems either for (1) explicit user queries or (2) implicit user queries for proactive action suggestions, see Figure \ref{fig:teaser}. Both tasks work towards building real-world action recommendation systems for augmented and virtual reality systems.

\subsection{Task 1: Explicit Query-to-Action}

Given the query from the \ours{} dataset and the corresponding context from the Ego4D dataset, the task is to predict the action that the system should call on in order handle the user query. The \ours{} dataset provides human annotations for six kinds of actions, thus making this a classification task with $C=6$ classes. Formally, the task is to approximate the function $f: (c, q) \mapsto a$ where $a \in \{1, \ldots C\}$ is the action, $c$ is the context, and $q$ is the text describing the user query. The task can be solved in language-only mode by using the textual narration from the Ego4D dataset, or in multimodal mode by using the raw videos from the Ego4D dataset. Thus, the context $c$ can be either text (as the narrations), or a video stream, or a combination of multiple modalities. 
We report the baseline performance of this task with a prompt engineered Gemini Pro model used in zero-shot manner. The system prompt for this task is presented in the Supplementary Material. 


\subsection{Task 2: Implicit Query-to-Action}

For increased autonomy of the AI assistant and easier interfacing for users of AR and VR systems, we further propose a new benchmark task to evaluate a system's capability to make action suggestions without an explicit user query. Instead, the model only receives a signal of intent from the user, for example the press of an action button or the invocation of a hot word without an explicit query that specifies the user's intent. The present dataset inherently contains such intent signals - which are the timesteps in the Ego4D data for which the \ours{} dataset contains verified \texttt{sensible} suggestions.

\begin{table}[]
    \centering
    \caption{
    Baseline results for the \ours{} benchmark tasks. All models use language-based input through the narrations provided by the Ego4D dataset annotations. The Gemini model is used in zero-shot manner. 
    \textbf{Left}: shows the Explicit-Query-to-Action task where classification accuracy is reported. The constant model always predicts the most frequent action in the training dataset. 
    \textbf{Right}: shows the Implicit Query-to-Action task where negative log likelihood is reported, see main text for explanation. The first random model randomly predicts one of the top-500 (query, action) pairs, the second random model chooses from all (query, action) pairs in the training dataset. 
    }
    \vspace{0.2cm}
    \begin{tabular}{lccc}
    \toprule[1.5pt]
    Model       & Train  & Val & Test \\
    \midrule
    Gemini Pro  & 55.95\% & 54.43\% & 63.57\% \\
    Constant    & 42.75\% & 42.75\% & 42.75\% \\
    \\
    \bottomrule[1.5pt]
    \end{tabular}
    \quad
    \begin{tabular}{lccc}
    \toprule[1.5pt]
    Model           & Train  & Val & Test \\
    \midrule
    Gemini Pro      & -43.43 & -43.46 & -42.50 \\
    Random (top)    & -44.77 & -45.07 & -44.80 \\
    Random (all)    & -53.68 & -53.97 & -53.39 \\
    \bottomrule[1.5pt]
    \end{tabular}
    \label{tbl:benchmark-performance}
\end{table}

The input is the context at a given point in time from the Ego4D dataset where the time is taken from the \ours{} dataset, filtered as described in Section \ref{ss:human-annotation-study}. As with the previous task, the context can be ingested either in language form, as narrations from the Ego4D dataset annotations, or in raw video form. We present baseline results for the language-based narrations as input. 
The output for this task can be an action suggestion, as shown in Figure \ref{fig:teaser}. However, it is evident that all necessary information about such an action suggestion is also contained in the (query, action) pair that is provided in the \ours{} dataset. As such, we propose to solve this task by learning the function $f: c \mapsto (q, a)$ where $c$ is the context from the Ego4D dataset, and $(q,a)$ is the (query, action) tuple. 
As this is an open-ended task with the final output being in natural language, we propose the use of the negative log-likelihood of the language model's output on the (query, action) pair from the \ours{} dataset, given the Ego4D context as input. We report the performance of a baseline LLM model on text-based narrations as context input, and provide two naive baseline methods for comparison, see Table \ref{tbl:benchmark-performance}. The system prompt for the LLM on this task is presented in the Supplementary Material.

\section{Discussion and Limitations}

\paragraph{Context only as textual narrations} 
We generated the presented dataset based only on textual narrations from the Ego4D dataset that were provided by human annotators. Using a the few-shot learning ability of foundation models would, at the present time, be too computationally expensive on video data directly. However, it is conceivable to pass one, or a few, images from the video stream into the model, along with the textual narrations. It may also be advantageous to train a video-to-text model directly or fine-tune an existing model using our proposed dataset. Experiments using multimodal LLMs on our proposed benchmark tasks remain to be explored.

\paragraph{Efficient ML systems}
Our proposed experimental baselines use either naive methods or a state-of-the-art LLM that is too large to be deployed on AR/VR devices. We encourage future work to explore the tradeoffs between performance on the proposed tasks and the efficiency of the suggestion model. Novel efficient architectures for sequence modeling \cite{botev2024recurrentgemma,dao2024transformers,gu2023mamba} may provide a path towards efficient AI assistants running on-device in resource-constrained environments such as those faced by AR/VR systems.

\paragraph{Moving beyond human annotations} 
Despite in our approach we use LLMs to create the dataset through prompt engineering on the narration of videos, we still require a certain level of human annotation to evaluate the quality of the dataset. This is inline with current recommendations that test the limits of how far can synthetic user experiences go \cite{jie2014synthetic}. It remains to be explored if new advances in self-training LLMs based on automated scalar feedback \cite{singh2023beyond} or self-consistency \cite{huang2022large} can be applied to our dataset to improve the performance of LLMs on our proposed tasks.

\paragraph{Multi-turn suggestions and bespoke UI} 
The development of personalized action recommendation systems in egocentric video presents a unique challenge in the design of user interfaces (UI). Traditional Assistants rely on queries by user, often optimized for general use, may not be suitable for presenting contextually relevant suggestions unless users start doing multi-turn interactions. This necessitates the exploration of shortcuts and bespoke UI designs that can seamlessly integrate with the user's context. In our research we propose implicit queries that can actually reduce the number of multi-turn queries or UI interactions needed. 

\paragraph{Advanced LLM reasoning techniques.} The creation of our \ours{} dataset aligns with and could benefit from advancements in Large Language Model (LLM) reasoning techniques, specifically Chain-of-Thought (CoT) \cite{wei2022chain}, Tree-of-Thought (ToT) \cite{long2023large,yao2024tree}, and self-reflection \cite{ji2023towards}. These techniques hold the potential to enhance both the generation and evaluation of action suggestions, moving us closer to truly personalized AI assistants.
CoT prompting encourages LLMs to generate intermediate reasoning steps before reaching a conclusion. This approach can be applied to action suggestions by prompting the LLM to explicitly consider the user's context, goals, and preferences before recommending an action. For example, instead of directly suggesting ``Turn on the lights'', the LLM might first reason about the time of day, the user's location, and their recent activities. This could lead to more nuanced suggestions like, ``It's getting dark in the kitchen, would you like me to turn on the lights?''.
ToT extends CoT by allowing the LLM to explore multiple reasoning paths in parallel. This could be beneficial for generating a wider range of action suggestions and evaluating their potential impact on the user. For instance, the LLM could consider different options for completing a task, weigh their pros and cons, and present the most suitable one to the user.
Self-reflection enables LLMs to evaluate their own outputs and identify potential errors or biases. In the context of action suggestions, this could involve the LLM assessing the confidence of its recommendations and providing explanations to the user. This could increase user trust and allow them to understand the reasoning behind the suggestions.
%
As LLM reasoning techniques advance, they will also open up new research avenues, such as developing LLM-based agents that can learn user preferences and adapt their suggestions over time. CoT and ToT prompting could be used in real-time to refine the LLM-generated action suggestions in \ours{}, making them more contextually relevant, time bonded, and personalized.

\section{Conclusion and Broader Impacts}

In this work, we have introduced \ours{}, a novel dataset that expands upon the existing Ego4D dataset by incorporating context-aware personal action recommendation annotations. By leveraging a two-stage annotation process combining automated suggestions from a large language model (Gemini Pro) and human evaluation, we have ensured the quality, relevance, and usefulness of these recommendations. Our comprehensive human evaluation not only validates the efficacy of the LLM-generated suggestions but also reveals insights into the nuances of user preferences in real-world scenarios, for example proposing a difference between implicit and explicit types of queries. Through this dataset, we aim to empower researchers and developers to build intelligent assistants capable of anticipating user needs and proactively offering personalized action suggestions, ultimately enhancing the user experience in egocentric video applications.  

Our dataset also is free of personally identifiable information and given the very tailored prompt engineering eliminates the appearance of offensive content. Both aspects are also further enhanced by the reliance on the original Ego4D dataset. The annotations in \ours{} will support future research on a variety of tasks, such as intent to action mapping, personalized suggestion learning, and user modeling.  We believe that the release of this dataset will significantly advance the field of proactive AI assistance in egocentric video and contribute to the development of more intelligent and intuitive user experiences.


\clearpage
\bibliography{name}{}
\bibliographystyle{plain}



\clearpage
\section*{Checklist}


\begin{enumerate}

\item For all authors...
\begin{enumerate}
  \item Do the main claims made in the abstract and introduction accurately reflect the paper's contributions and scope?
    \answerYes{}
  \item Did you describe the limitations of your work?
    \answerYes{Limitations are discussed in Section 5.}
  \item Did you discuss any potential negative societal impacts of your work?
    \answerYes{The broader impacts of our work are discussed in Section 7.}
  \item Have you read the ethics review guidelines and ensured that your paper conforms to them?
    \answerYes{}
\end{enumerate}

\item If you are including theoretical results...
\begin{enumerate}
  \item Did you state the full set of assumptions of all theoretical results?
    \answerNA{No theoretical results are included.}
	\item Did you include complete proofs of all theoretical results?
    \answerNA{No theoretical results are included.}
\end{enumerate}

\item If you ran experiments (e.g. for benchmarks)...
\begin{enumerate}
  \item Did you include the code, data, and instructions needed to reproduce the main experimental results (either in the supplemental material or as a URL)?
    \answerYes{Code, data and instructions for reproducing the experimental results can be found on our project page.}
  \item Did you specify all the training details (e.g., data splits, hyperparameters, how they were chosen)?
    \answerYes{}
	\item Did you report error bars (e.g., with respect to the random seed after running experiments multiple times)?
    \answerYes{}
	\item Did you include the total amount of compute and the type of resources used (e.g., type of GPUs, internal cluster, or cloud provider)?
    \answerYes{}
\end{enumerate}

\item If you are using existing assets (e.g., code, data, models) or curating/releasing new assets...
\begin{enumerate}
  \item If your work uses existing assets, did you cite the creators?
    \answerYes{Our work builds on and contributes to the Ego4D dataset which we have cited.}
  \item Did you mention the license of the assets?
    \answerYes{}
  \item Did you include any new assets either in the supplemental material or as a URL?
    \answerYes{}
  \item Did you discuss whether and how consent was obtained from people whose data you're using/curating?
    \answerYes{We have provided detailed information about the instructions and information given to annotators recruited for this work.}
  \item Did you discuss whether the data you are using/curating contains personally identifiable information or offensive content?
    \answerYes{}
\end{enumerate}

\item If you used crowdsourcing or conducted research with human subjects...
\begin{enumerate}
  \item Did you include the full text of instructions given to participants and screenshots, if applicable?
    \answerYes{Yes, we have included details of the instructions and screenshots of the annotation interface.}
  \item Did you describe any potential participant risks, with links to Institutional Review Board (IRB) approvals, if applicable?
    \answerNA{Risks associated with the annotation task were deemed very minimal.}
  \item Did you include the estimated hourly wage paid to participants and the total amount spent on participant compensation?
    \answerYes{Details of the tasks and compensation are included in Section 3.6 "Participants".}
\end{enumerate}

\end{enumerate}


\clearpage
\appendix
\section{Appendix}

\subsection{Dataset Availability}
The dataset is available on the \ours{} GitHub repository on: \url{https://github.com/parse-ego4d/parse-ego4d.github.io/tree/main/dataset/}.

\subsection{Human Annotation Demographics}

\begin{figure*}[h!]
    \centering
    \includegraphics[width=\textwidth]{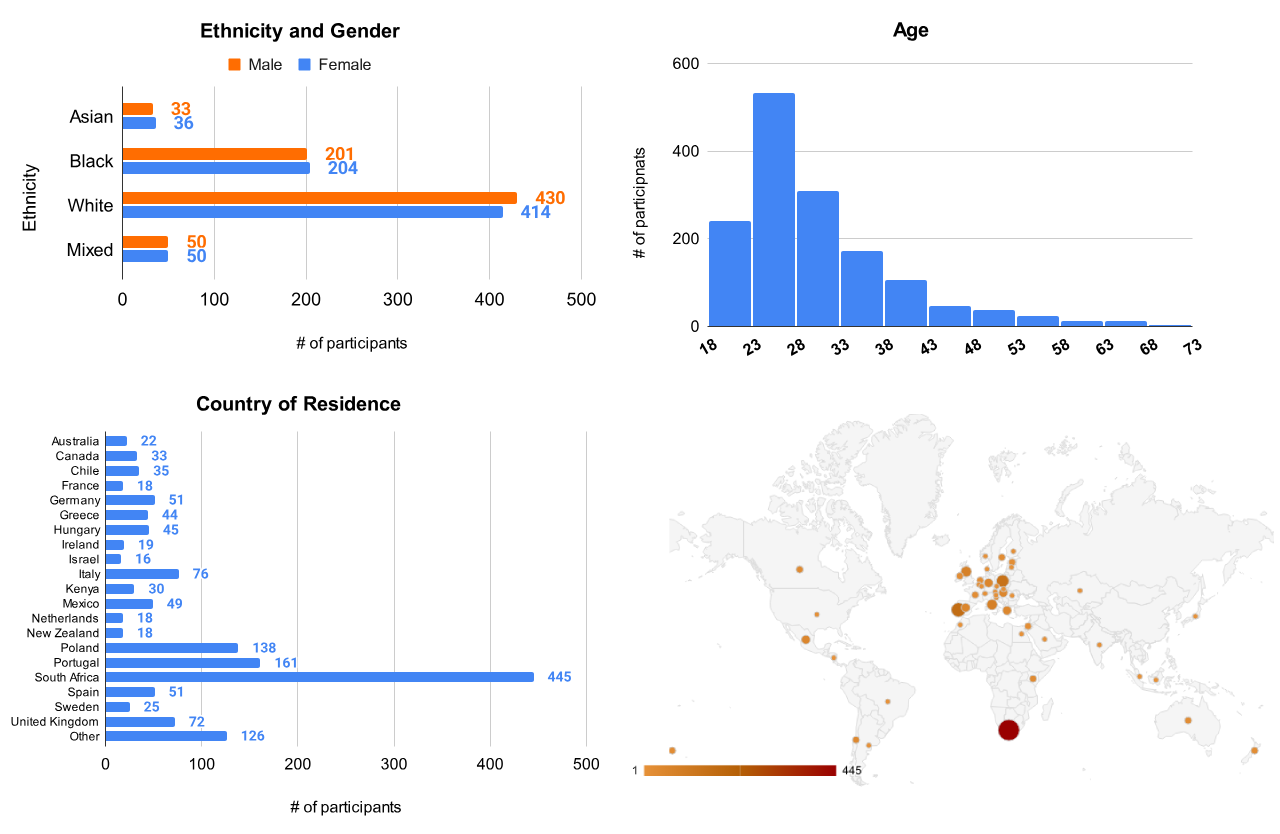}
    \caption{A demographic breakdown of our participants in the annotation study, including ethnicity, gender, and age. Countries with fewer than 15 participants are listed in "Other".}
    \label{fig:demographics}
\end{figure*}

A visualization of the demographics from our human annotation study is presented in Figure \ref{fig:demographics}.

\subsection{Annotation Metrics}

The human annotations are used to filter the suggestions in \ours{} so that samples above a certain mean rating for each question are accepted. Table \ref{table:metrics} shows an overview of how many samples are accepted at different mean ratings.

\begin{table*}[h!]
\centering
\begin{tabular}{l|c|c}
\toprule
Filter & Percentage & Number of Suggestions \\
\midrule
All samples & 100.00\% & 18,360 \\
\midrule
\texttt{sensible} $\ge$ 3 & 78.10\% & 14,340 \\
\texttt{sensible} $\ge$ 3.5 & 63.10\% & 11,586 \\
\texttt{sensible} $\ge$ 4 & 58.31\% & 10,705 \\
\midrule
\texttt{correct} $\ge$ 3 & 74.56\% & 13,689 \\
\texttt{correct} $\ge$ 3.5 & 59.54\% & 10,932 \\
\texttt{correct} $\ge$ 4 & 54.80\% & 10,061 \\
\midrule
\texttt{implicit} $\ge$ 3 & 59.38\% & 10,903 \\
\texttt{implicit} $\ge$ 3.5 & 37.61\% & 6,905 \\
\texttt{implicit} $\ge$ 4 & 33.26\% & 6,107 \\
\midrule
\{\texttt{sensible}, \texttt{correct}\} $\ge$ 3 & 65.00\% & 11,934 \\
\{\texttt{sensible}, \texttt{correct}\} $\ge$ 3.5 & 47.17\% & 8,660 \\
\{\texttt{sensible}, \texttt{correct}\} $\ge$ 4 & 42.32\% & 7,770 \\
\midrule
\{\texttt{sensible}, \texttt{correct}, \texttt{implicit}\} $\ge$ 3 & 48.22\% & 8,854 \\
\{\texttt{sensible}, \texttt{correct}, \texttt{implicit}\} $\ge$ 3.5 & 27.65\% & 5,076 \\
\{\texttt{sensible}, \texttt{correct}, \texttt{implicit}\} $\ge$ 4 & 24.02\% & 4,410 \\
\bottomrule
\end{tabular}
\caption{Number of suggestions in \ours{} above a mean rating for different metrics. The filter \texttt{\{sensible,correct\}} is applied for Task 1, whereas the \texttt{\{sensible,correct,implicit\}} filter is applied for Task 2.}
\label{table:metrics}
\end{table*}


\subsection{Annotation Interface Screenshots}

\begin{figure*}[h!]
    \centering
    \begin{subfigure}[b]{0.475\textwidth}
        \centering
        \includegraphics[height=2.5in]{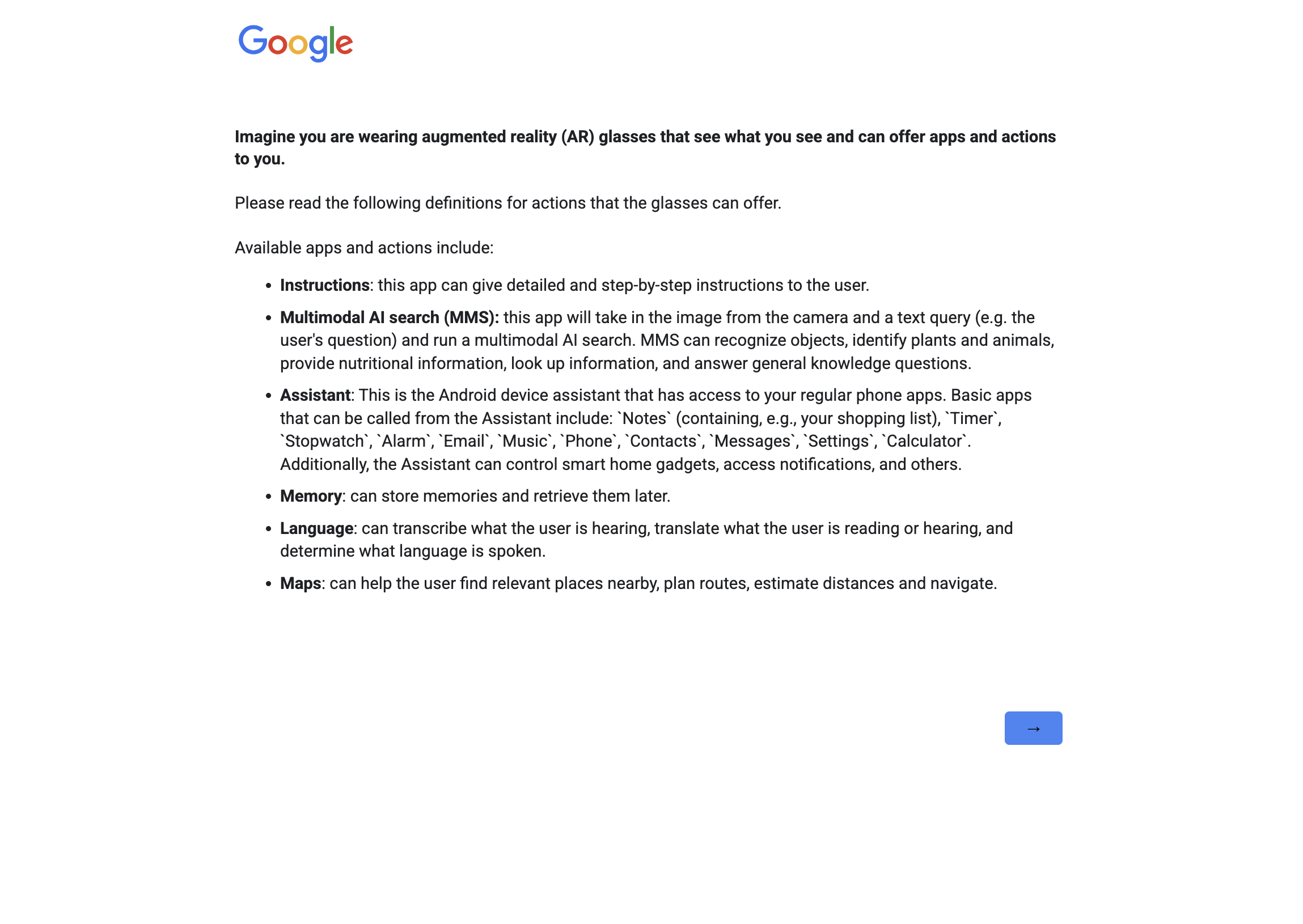}
        \caption{Introductory instructions.}
    \end{subfigure}%
    \hfill
    \begin{subfigure}[b]{0.475\textwidth}
        \centering
        \includegraphics[height=2.5in]{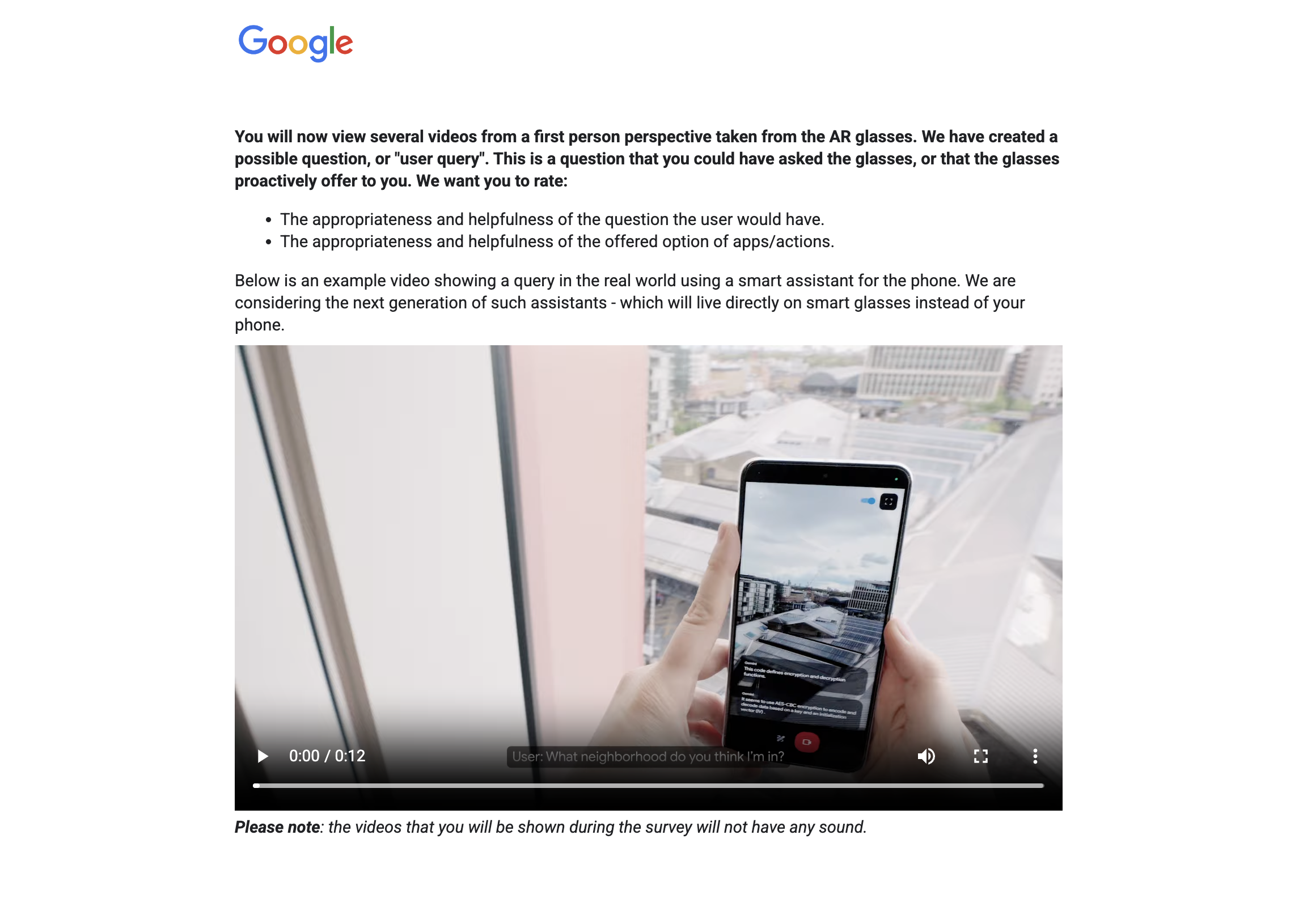}
        \caption{Example video.}
    \end{subfigure}
    \vskip\baselineskip
    \begin{subfigure}[b]{0.475\textwidth}
        \centering
        \includegraphics[height=2.5in]{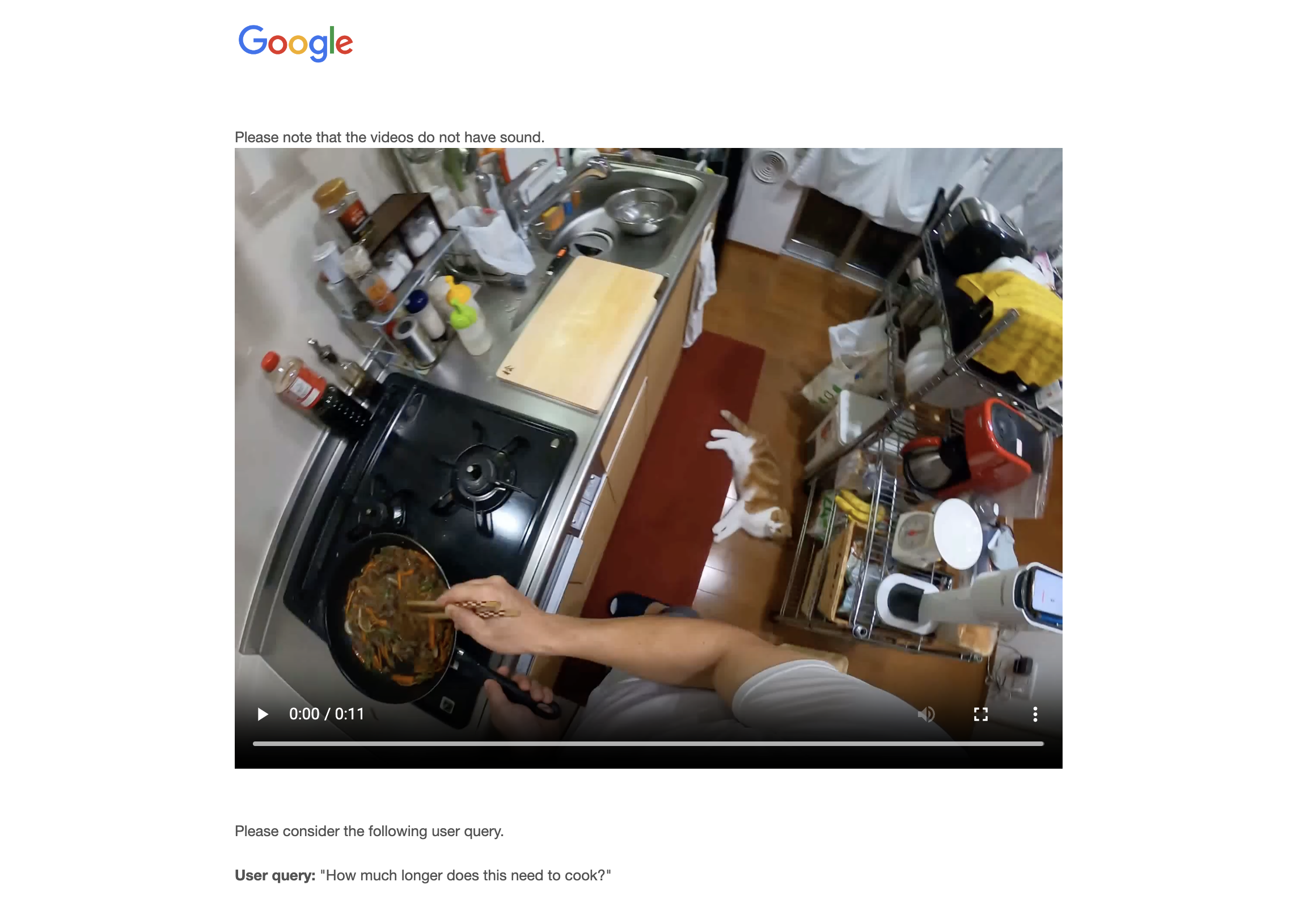}
        \caption{Task video.}
    \end{subfigure}
    \hfill
    \begin{subfigure}[b]{0.475\textwidth}
        \centering
        \includegraphics[height=2.5in]{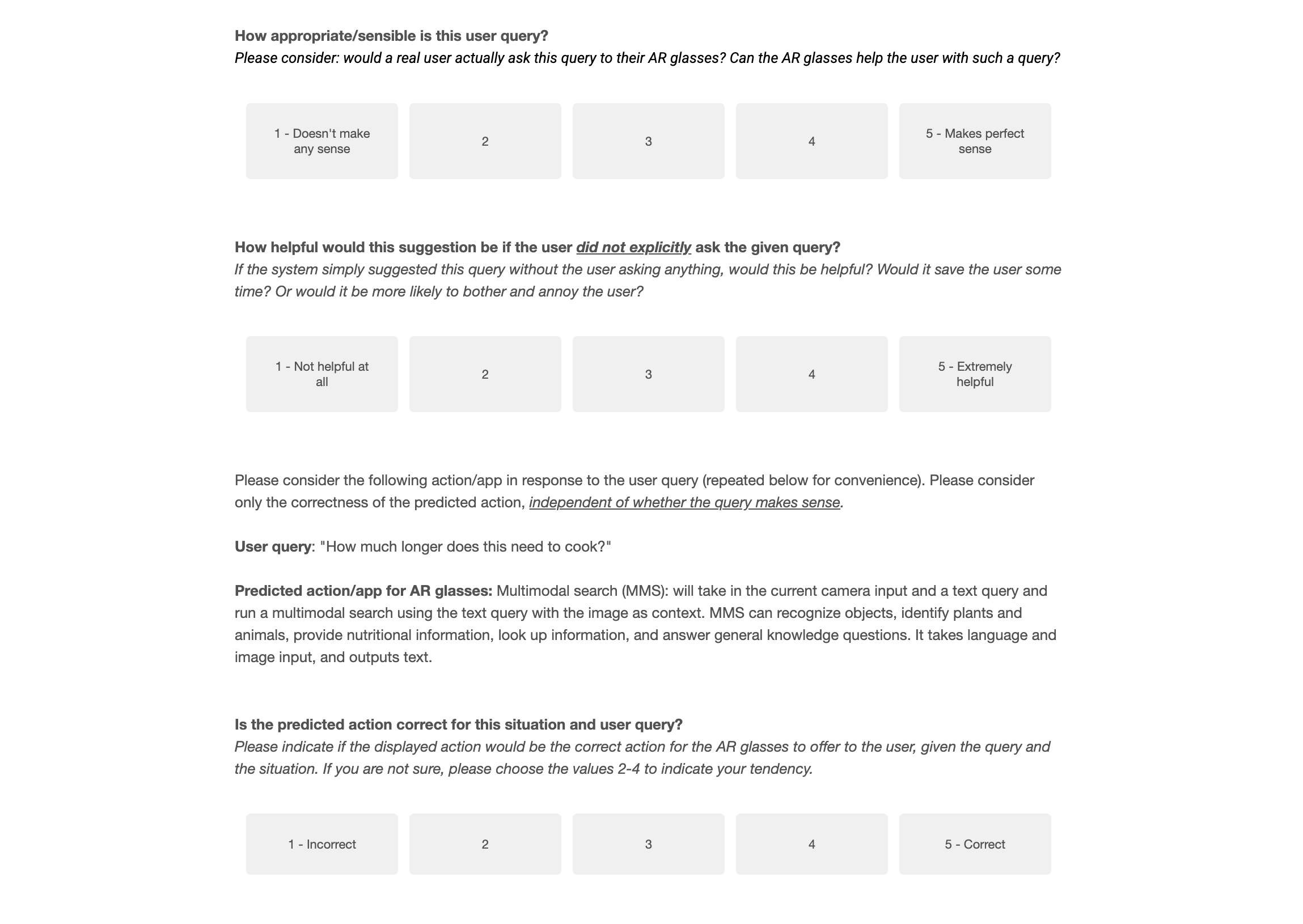}
        \caption{Task questions.}
    \end{subfigure}
    \caption{Screenshots of the human annotation task.}
    \label{fig:screenshot-annotation}
\end{figure*}

The human annotation study was run using Prolific, with participants filling out the survey on Qualtrics. The survey design is illustrated in Figure \ref{fig:survey-design} and Figure \ref{fig:screenshot-annotation} shows screenshots of the survey that human participants filled out.

\end{document}